\documentclass{article}

\usepackage{hyperref}

\usepackage[accepted]{icml2021_oppo}%[accepted]

\usepackage{microtype}
\usepackage{graphicx}
\usepackage{subfigure}
\usepackage{booktabs} % for professional tables
\usepackage{epsfig}
\usepackage{graphicx}
\usepackage{amsmath}
\usepackage{amssymb}
\usepackage{url}
\usepackage{enumitem}
\usepackage{wrapfig}
\usepackage{booktabs}       % professional-quality tables
\usepackage{algorithm}
\usepackage{algpseudocode}
\usepackage{multirow}
\usepackage{xcolor}
\usepackage{amsthm}
\usepackage{marginnote}

\icmltitlerunning{Mitigating deep double descent by concatenating inputs}

\begin{document}

\twocolumn[
\icmltitle{Mitigating deep double descent by concatenating inputs}



\icmlsetsymbol{equal}{*}

\begin{icmlauthorlist}
\icmlauthor{John Chen}{rice}
\icmlauthor{Qihan Wang}{rice}
\icmlauthor{Anastasios Kyrillidis}{rice}
% \icmlauthor{Iaesut Saoeu}{ed}
% \icmlauthor{Fiuea Rrrr}{to}
% \icmlauthor{Tateu H.~Yasehe}{ed,to,goo}
% \icmlauthor{Aaoeu Iasoh}{goo}
% \icmlauthor{Buiui Eueu}{ed}
% \icmlauthor{Aeuia Zzzz}{ed}
% \icmlauthor{Bieea C.~Yyyy}{to,goo}
% \icmlauthor{Teoau Xxxx}{ed}
% \icmlauthor{Eee Pppp}{ed}
\end{icmlauthorlist}

\icmlaffiliation{rice}{Department of Computer Science, Rice University, Houston, USA}
% \icmlaffiliation{to}{Department of Computation, University of Torontoland, Torontoland, Canada}
% \icmlaffiliation{goo}{Googol ShallowMind, New London, Michigan, USA}
% \icmlaffiliation{ed}{School of Computation, University of Edenborrow, Edenborrow, United Kingdom}

\icmlcorrespondingauthor{John Chen}{johnchen@rice.edu}
% \icmlcorrespondingauthor{Eee Pppp}{ep@eden.co.uk}

\icmlkeywords{Machine Learning, ICML}

\vskip 0.3in
]



\printAffiliationsAndNotice{\icmlEqualContribution} % otherwise use the standard text.
\begin{abstract}
The double descent curve is one of the most intriguing properties of deep neural networks. 
It contrasts the classical bias-variance curve with the behavior of modern neural networks, occurring where the number of samples nears the number of parameters. 
In this work, we explore the connection between the double descent phenomena and the number of samples in the deep neural network setting.
In particular, we propose a construction which augments the existing dataset by artificially increasing the number of samples. 
This construction empirically mitigates the double descent curve in this setting. 
We reproduce existing work on deep double descent, and observe a smooth descent into the overparameterized region for our construction. 
This occurs both with respect to the model size, and with respect to the number epochs.
\end{abstract}

\section{Introduction}
Underparameterization and overparameterization are at the heart of understanding modern neural networks. 
The traditional notion of underparameterization and overparameterization led to the classic U-shaped generalization error curve \citep{hastie2001the, gemen1992neural}, where generalization would worsen when the model had either too few (underparameterized) or too many parameters (overparameterized). 
Correspondingly, it was expected that an underparameterized model would underfit and fail to identify more complex and informative patterns, and an overparameterized model would overfit and identify non-informative patterns. 

This view no longer holds for modern neural networks. 
It is widely accepted that neural networks are vastly overparameterized, yet generalize well. 
There is strong evidence that increasing the number of parameters leads to better generalization \citep{zagoruyko2016wide, huang2017dense, larsson2016ultra}, and models are often trained to achieve zero training loss \citep{salak2017deep}, while still improving in generalization error, whereas the traditional view would suggest overfitting. 

To bridge the gap, \cite{belkin2018reconciling} proposed the double descent curve, where the underparameterized region follows the U-shaped curve, and the overparameterized region smoothly decreases in generalization error, as the number of parameters increases further. 
This results in a peak in generalization error, where a fewer number of samples would counter-intuitively decrease the error. 
There has been extensive experimental evidence of the double descent curve in deep learning \citep{nakkiran2019deepdouble, yang2020rethinking}, as well as in models such as random forests, and one layer neural networks \citep{belkin2018reconciling, ba2019generalization}. 

One recurring theme in the definition of overparameterization and underparameterization lies in the number of neural network parameters relative to the number of samples \citep{belkin2018reconciling, nakkiran2019deepdouble, ba2019generalization,bibas2019new, muthukumar2019harmless,hastie2019surprises}.
On a high level, a greater number of parameters than samples is generally considered overparameterization, and fewer is considered underparameterization. 

However, this leads to the question ``\emph{What is a sample?}''
In this paper, we revisit the fundamental underpinnings of overparameterization and underparameterization, and stress test when it means to be overparameterized or underparameterized, through extensive experiments of a cleverly constructed input.
We artificially augment existing datasets by simply stacking every combination of inputs, and show the mitigation of the double descent curve in the deep neural network setting. 
We humbly hypothesize that in deep neural networks we can, perhaps, artificially increase the number of samples without increasing the information contained in the dataset, and by implicitly changing the classification pipeline mitigate the double descent curve. %\textcolor{magenta}{State the hypothesis: that we can artificially extend the number of samples by not increasing the information contained in the dataset, but implicitly changing the process of classification. This further signifies that maybe the double descent curve is an artifact of the current supervised training pipeline: one sample as input at a time. Can we state such a hypothesis without being too "cocky" since we have no theory to back this up?}
In particular, the narrative of our paper obeys the following:  \vspace{-0.1cm}
\begin{itemize}%[leftmargin=*]
    \item We propose a simple construction to artificially augment existing datasets of size $O(n)$ by stacking inputs to produce a dataset of size $O(n^2)$. \vspace{-0.1cm}
    \item We demonstrate that the construction has no impact on the double descent curve in the linear regression case. \vspace{-0.1cm}
    \item We show experimentally that those results on double descent curve do not extend to the case of neural networks. \vspace{-0.1cm}
    %\item Lastly, we observe a shifting of the double descent curve when moving from cross entropy loss to binary cross entropy. 
\end{itemize}
Concretely, we reproduce recent landmark papers, presenting the differing behavior with respect to double descent.

\section{Related Works}
The double descent curve was proposed recently \citep{belkin2018reconciling}, where the authors define over/underparameterization as the proportion of parameters to samples, explained through model capacity. 
With more parameters in the overparameterized region, there is larger ``capacity'' (i.e., the model class contains more candidates), and thus may contain better, simpler models by Occam's Razor rule. 
The interpolation region is suggested to exist when the model capacity is capable of fitting the data nearly perfectly by overfitting on non-informative features, resulting in higher test error. 
%Experiments included a one-layer neural network, random forests, and others.

Double descent is also observed in deep neural networks \citep{nakkiran2019deepdouble}, in addition to  epoch-wise double descent. 
Experimentation is amplified by label noise.
With the observation of unimodel variance \citep{neal2018modern}, \cite{yang2020rethinking} decomposes the risk into bias and variance, and posits that double descent arises due to the bell-shaped variance curve rising faster than the bias decreases.

There is substantial theoretical work on double descent, particularly in the least squares regression setting.
\cite{advani2017high} analyses this linear setting and proves the existence of the interpolation region, where the number of parameters equals the number of samples in the asymptotic limit.
Another work \cite{hastie2019surprises} proves that regularization reduces the peak in the interpolation region. 
\cite{belkin2019two} requires only finite samples, where the features and target be jointly Gaussian. 
Other papers with similar setup include \citep{muthukumar2019harmless, bibas2019new, mitra2019understanding, mei2019generalization,ba2019generalization,nakkiran2019data,bartlett2019benign,chen2020multiple}. 
%Additionally, there is supporting evidence of double descent in the sample-wise perspective. %\citep{nakkiran2020optimal}.

% \cite{ba2019generalization} analyses the least squares regression setting for two layer linear neural networks in the asymptotic setting, where the double descent curve is present when only the second layer is optimized. 
% There is also work in proving that optimally tuned $\ell_2$-norm regularization mitigates the double descent curve for certain linear regression models with isotropic data distribution \citep{nakkiran2019data}. 
% This setting has also been studied with respect to the variance in the parameter space \citep{bartlett2019benign}. 
% Multiple descent has also been studied, including showing in the linear regression setting that multiple descent curves can be directly designed by the user \citep{chen2020multiple}. 
% Additionally, there is supporting evidence of double descent in the sample-wise perspective \citep{nakkiran2020optimal}. 

% There is other work in this area, including studying the double descent curve for least squares in random feature models \citep{belkin2019models, dascoli2020triple, ghorbani2019linearized}, using Neural Tangent Kernel\citep{Geiger_2020}, characterizing double descent in non-linear settings \citep{caron2020finite}, kernel learning \citep{belkin2018understand, liang2019multiple}, and connecting to other fields \citep{Geiger_2019}. 
% Lastly, we note here that, in the deep neural network setting, models can be trained to zero training loss even with random labels \citep{zhang2016understanding}.

\section{Methods}

We introduce the \emph{concatenated inputs construction}. 
This way the size of a dataset can be artificially, but non-trivially, increased.
This construction can be applied both to the regression setting and the classification setting. 
In linear regression, for given input pairs, $(x_1, y_1)$, $(x_2, y_2)$, an augmented dataset can be constructed: \vspace{-0.1cm}
\begin{align*}
    \Big\{\left([x_1, x_1], \tfrac{y_1 + y_1}{2} \right), \left([x_1, x_2], \tfrac{y_1 + y_2}{2}\right), \\
    \left([x_2, x_1], \tfrac{y_2 + y_1}{2}\right), \left([x_2, x_2], \tfrac{y_2 + y_2}{2}\right) \Big\}, \vspace{-0.1cm}
\end{align*} 
where $[\alpha, \beta]$ represents concatenation of the input $\alpha, \beta$.
In the setting of classification, the process is identical, where the targets are produced by element-wise addition and then averaged to sum to 1. 
The averaging is not strictly necessary even in the deep neural network classification case, where the binary cross entropy loss can be used instead of cross entropy.
% However, we adhere to the convention that probabilities sum to 1, and demonstrate further that using the binary cross entropy loss produces interesting behavior with respect to the cross entropy loss. 
For test data, we concatenate the same input with itself, and the target is the original target. This way a dataset of size $O(n)$ is artificially augmented to size $O(n^2)$.

% \begin{figure}
%   \centering
%   \begin{subfigure}[b]{0.49\linewidth}
%     \includegraphics[width=\linewidth]{LinRegPlots/oneHot.png}
%     \label{fig:oneHotLinReg}
%   \end{subfigure} \hspace{-0.1cm}
%   \begin{subfigure}[b]{0.49\linewidth}
%     \includegraphics[width=\linewidth]{LinRegPlots/twoHot.png}
%     \label{fig:twoHotLinReg}
%   \end{subfigure} 
%   \caption{ 
%   %\textcolor{magenta}{Please increase the font size of axes, ass well as the lenges on the axes} 
%   Left: The standard case. Right: The concatenated inputs construction. Plots of the Test MSE verses number of samples (pre-concatenation) for min-norm ridgeless regression, where $d=30$. Following \cite{nakkiran2019data}, inputs are drawn $x \sim \mathcal{N}(0, I_d)$, target $y = \theta x + \mathcal{N}(0, \sigma^2)$, where $\theta$ are the parameters, $||\theta||_2 = 1$, $\sigma = 0.1$. $\hat{\theta} = X^{\dagger} y$.}
%   \label{fig:linReg}
% \end{figure}

Our reasons for the concatenated inputs construction are as follows:
$i)$ there is limited injection of information or semantic meaning;
$ii)$ the number of samples is significantly increased.
For the purposes of understanding underparameterization, overparameterization and double descent, such a construction tries to isolate the definition of a sample. %If underparameterization, overparameterization and the double descent curve are sensitive to such a construction, it will show the truly uncanny behavior of neural networks to respond to only the number of samples without additional information. 

\section{Results}
In this section, we reproduce settings from benchmark double descent papers, add the concatenated inputs construction and analyze the findings. 
%In particular, we begin with the linear regression, move to one hidden layer feedforward neural networks, and then deep neural networks, in both the model parameter-wise and epoch-wise double descent. 
\begin{figure}[ht]
  \centering
  \begin{minipage}[b]{0.49\linewidth}
    \includegraphics[width=\linewidth]{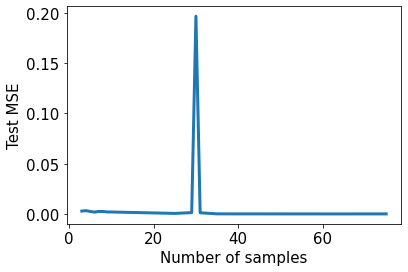}
    \label{fig:oneHotLinReg}
  \end{minipage} \hspace{-0.1cm}
  \begin{minipage}[b]{0.49\linewidth}
    \includegraphics[width=\linewidth]{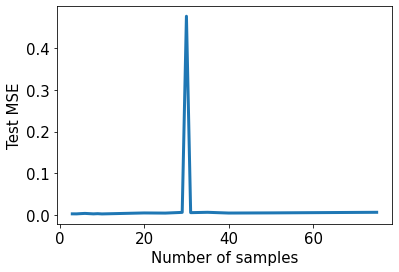}
    \label{fig:twoHotLinReg}
  \end{minipage} 
  \vspace{-0.6cm}
  \caption{ 
  %\textcolor{magenta}{Please increase the font size of axes, ass well as the lenges on the axes} 
  Left: The standard case. Right: The concatenated inputs construction. Plots of the Test MSE versus the number of samples (pre-concatenation) for min-norm ridgeless regression, where $d=30$. Following \cite{nakkiran2019data}, inputs are drawn $x \sim \mathcal{N}(0, I_d)$, target $y = \theta x + \mathcal{N}(0, \sigma^2)$, where $\theta$ are the parameters, $||\theta||_2 = 1$, $\sigma = 0.1$. $\hat{\theta} = X^{\dagger} y$. As expected, the concatenated inputs construction does not affect the double descent curve, and the peak occurs in the exact same location.} \vspace{-0.1cm}
  \label{fig:linReg}
\end{figure}

\begin{figure}
  \centering
  %\begin{minipage}[b]%{1\textwidth}
    \includegraphics[width=\linewidth]{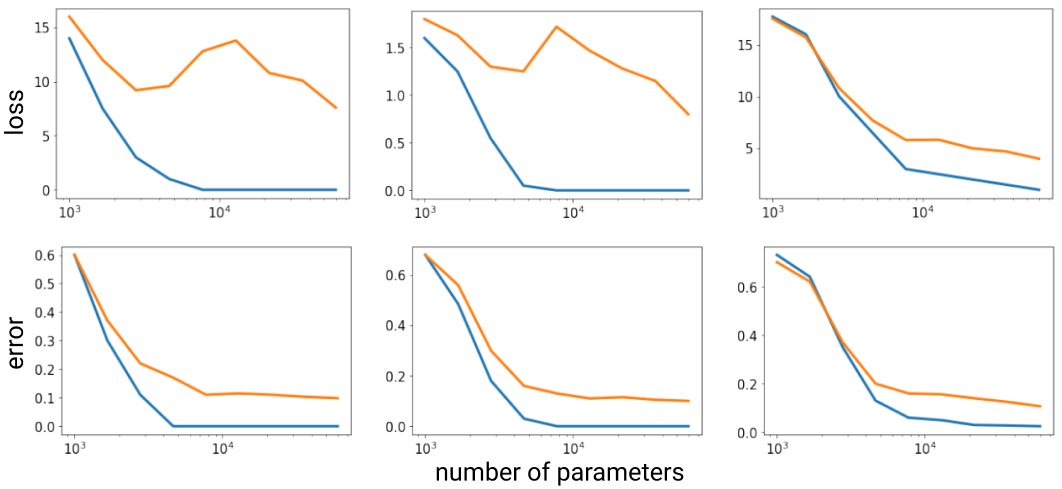}
    \label{fig:belkinPlots} \vspace{-0.6cm}
  %\end{subfigure}
%   \begin{subfigure}[b]{0.32\linewidth}
%     \includegraphics[width=\linewidth]{BelkinPaperPlots/oneHotCELoss.png}
%     \label{fig:oneHotBelkinLoss}
%   \end{subfigure} \hspace{-0.1cm}
%   \begin{subfigure}[b]{0.32\linewidth}
%     \includegraphics[width=\linewidth]{BelkinPaperPlots/oneHotStackedCELoss.png}
%     \label{fig:oneHotStackedCEBelkinLoss}
%   \end{subfigure} \hspace{-0.1cm}
%   \begin{subfigure}[b]{0.32\linewidth}
%     \includegraphics[width=\linewidth]{BelkinPaperPlots/twoHotStackedMixupCELoss.png}
%     \label{fig:twoHotMixupCEBelkinLoss}
%   \end{subfigure} \hspace{-0.1cm}
%   \begin{subfigure}[b]{0.32\linewidth}
%     \includegraphics[width=\linewidth]{BelkinPaperPlots/oneHotCEError.png}
%     \label{fig:oneHotBelkinError}
%   \end{subfigure} \hspace{-0.1cm}
%   \begin{subfigure}[b]{0.32\linewidth}
%     \includegraphics[width=\linewidth]{BelkinPaperPlots/oneHotStackedCEError.png}
%     \label{fig:oneHotStackedCEBelkinError}
%   \end{subfigure} \hspace{-0.1cm}
%   \begin{subfigure}[b]{0.32\linewidth}
%     \includegraphics[width=\linewidth]{BelkinPaperPlots/twoHotStackedMixupCEError.png}
%     \label{fig:twoHotMixupCEBelkinError}
%   \end{subfigure} \hspace{-0.1cm}
  \caption{ 
  %Empirical results on adaptive learning rate methods.
  %\textcolor{magenta}{Presentation of results needs to be improved: increase all fontsize, increase with of curves to be more clear. Plots are not readable now.} 
  Top: Loss against number of parameters. Bottom: Error against number of parameters. In order from left to right: 1) Input is $32 \times 32$ image, label is one-hot vector; 2) Input is $64 \times 32$ image (two of the same image stacked on top of each other), label is one-hot vector; 3) Input is $64 \times 32$ image (two different $32 \times 32$ images stacked on top of each other), label is two-hot vector of values 0.5, 0.5. Cross Entropy loss. Orange/blue is validation/train. %\textcolor{magenta}{Tasos: Increase fontsize. What is the orange curve and what is the blue curve?} 
  } \vspace{-0.5cm}
  \label{fig:BelkinPlotNoBCE}
\end{figure}

\begin{figure*}
  \centering
  \includegraphics[width=\linewidth]{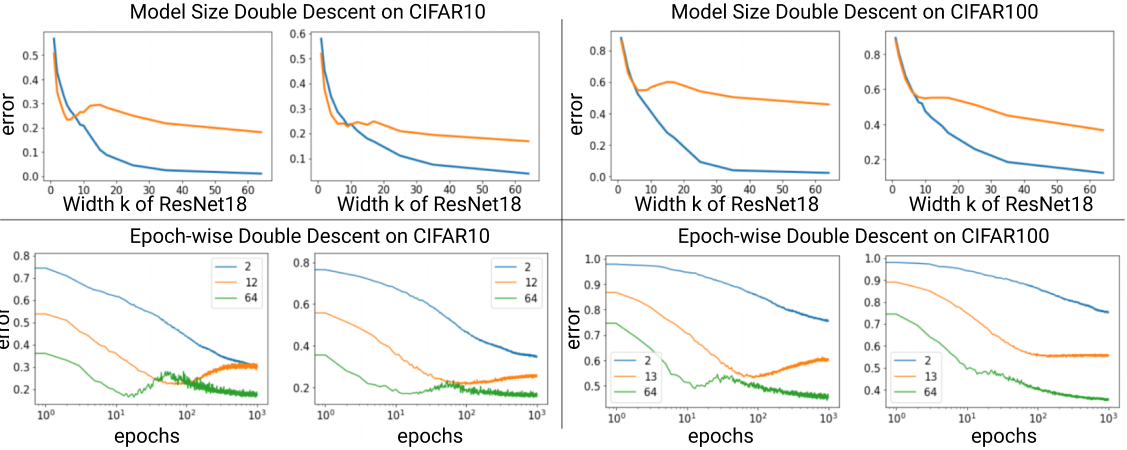} \vspace{-0.6cm}
  \caption{\small 
  %Empirical results on adaptive learning rate methods.
  %\textcolor{magenta}{Remove titles from the top, make fontsizes bigger, make curves thicker. Not sure if the top row is informatinve.} 
  Top Left: Model Size Double Descent on CIFAR10. Orange line is validation error, blue line is train error. Top Right: Model Size Double Descent on CIFAR100. Orange line is validation error, blue line is train error. Bottom Left: Epoch-wise Double Descent on CIFAR10. Bottom Right: Epoch-wise Double Descent on CIFAR100. For each subplot, Left: Standard one-hot vector setup, Right: Concatenated inputs construction. Bottom plots legends' represent the width k of the ResNet18. Models are ResNet18 architecture on CIFAR-10 and CIFAR-100 datasets. %\textcolor{magenta}{Tasos: what is the blue curves in the top row? Explain what we plot. Also do the numbers represent in the bottom-row plots. Try to include fontsize if possible.}
  }
  \label{fig:DLplot}
\end{figure*}

\subsection{Linear regression}
%\textcolor{magenta}{This section reads more like an appendiix section. Let's keep it for now.}
The linear regression setting has been a fruitful testbed for empirical work in double descent, as well as yielding substantial theoretical understanding. 
The concatenated inputs construction is applied similarly here, however with different motivation. 
Namely, we wish to motivate that the concatenated inputs construction is not expected to add any information and is therefore not expected to impact the double descent curve.

We reproduce the linear regression setting from \cite{nakkiran2019data}, given in Figure \ref{fig:linReg}. 
For the concatenated inputs construction, we first draw the number of samples before concatenation and construction of the augmented dataset. 
We observe that, by construction, the concatenated inputs construction does not affect the double descent curve, and the peak occurs in the exact same location.
We also remark here that it is not surprising, and it is not complicated to understand why from a theoretical perspective.

\subsection{One hidden layer feedforward neural network}

We train a feedforward neural network with one hidden ReLU layer on MNIST, reproducing the setup from \cite{belkin2018reconciling}.  We vary the number of parameters by changing the size of the hidden layer (see Figure \ref{fig:BelkinPlotNoBCE}). %The leftmost plots refer to the standard setup with a one-hot vector as target and Cross Entropy loss. 
%The middle plots refer to the standard setup with a one-hot vector as target and Binary Cross Entropy loss, where the double descent curve exists despite the change in loss function. %The rightmost plots refer to the two-hot vector setup with Binary Cross Entropy loss.

We observe in the upper-right plot in Figure \ref{fig:BelkinPlotNoBCE} the double descent curve is completely removed in the concatenated inputs construction relative to the other two settings. 
A smooth decrease in loss is observed, while there is a clear double descent in the other cases (upper-left/middle plots in Figure \ref{fig:BelkinPlotNoBCE}).
Furthermore, when concatenating each input only by itself (upper-middle plot in Figure (\ref{fig:BelkinPlotNoBCE})), the double descent curve is present almost exactly.% in this scenario. 
\emph{This provides evidence that the disappearance of the double descent is not due to the extra parameters, which originate from the larger sized inputs.}
In this setting, it appears that the behavior of under/overparameterization can be altered by artificially increasing the number of samples through concatenation. 

In addition, the model trained on MNIST and one-hot vectors can be concatenated with itself, with all other parameters being zero, to produce a model with two times the number of hidden units which can be applied to the concatenated inputs construction. 
We consider this setting in the context of a possible explanation of the interpolation region, where the number of parameters nears that of samples. 
Concretely, it is possible for a neural network with double the hidden units in the concatenated inputs construction to recover the double descent curve by learning two smaller, disconnected networks, where each of the smaller networks are the ones learned in the double descent peak of the standard, one-hot case. 
However, in practice while the network can do so, it does not appear to, which leads to the smooth descent in Figure \ref{fig:BelkinPlotNoBCE}.

\subsection{Deep Neural Networks}

\textbf{Model size double descent.}
We train a ResNet18 on CIFAR10 (top row in Figure \ref{fig:DLplot}), reproducing the setup from \cite{nakkiran2019deepdouble}. 
We vary the number of parameters in the neural network by changing the width $k$. %Again, we use the Cross Entropy loss for the standard setup and the Binary Cross Entropy loss for the two-hot vector. %The left plots refer to the standard setup with a one-hot vector as target and Cross Entropy loss. The right plots refer to the two-hot vector setup with Binary Cross Entropy loss. 
Immediately, we see that the double descent curve is relatively mitigated in the case of the concatenated inputs construction where we vary $k$ from $k=5$ to $20$. Here, the curve is much smoother, although not entirely smooth (see left (regular) and right (concatenated) subplots in the Top-Left cluster in Figure \ref{fig:DLplot}).
Notably, the concatenated inputs construction retains the test error across $k$, except in the interpolation region where it achieves significantly lower test error ($>5\%$). The conclusion is identical when error is plotted against parameters, since the concatenated inputs construction typically adds a very few parameters ($<5\%$).
We see this pattern repeated on CIFAR-100 (Figure \ref{fig:DLplot}, Top-right cluster).

\textbf{Epoch-wise double descent.}
We use the same setting, except we train models for an additional 600 epochs, totaling 1000. In Figure \ref{fig:DLplot}, Bottom row, we consider the ResNet18 architecture for CIFAR10 (Left cluster of plots) and CIFAR100 (Right cluster of plots). 
We plot test error against epochs. In the one hot setting (let us focus on the left plot of CIFAR10 case), as expected, we observe a U shape for medium sized models and a double descent for larger models \citep{nakkiran2019deepdouble}. 
For  concatenated inputs (Right plot of CIFAR10 case), a flat U shape and double descent is existent, but significantly mitigated. The mitigation allows a 5\% improvement at the end of training for medium sized models and a 10\% improvement in the middle of training for large models. We see similar results on CIFAR-100, where double descent disappears with concatenated inputs.

\subsection{Bias Variance decomposition}
In this section, we follow \cite{yang2020rethinking} and decompose the loss into bias and variance. Namely, let $\texttt{CE}$ denote the cross entropy loss, $T$ a random variable representing the training set, $\pi$ is the true one-hot label, $\bar{\pi}$ is the average log-probability after normalization, and $\hat{\pi}$ is the output of the neural network. Then,
\begin{equation*}
    \mathbb{E}_{T}[\texttt{CE}(\pi, \hat{\pi})] = D_{KL}(\pi||\bar{\pi}) + \mathbb{E}_{T}[D_{KL}(\bar{\pi}||\hat{\pi})]
\end{equation*}
where the first component is the bias and the second component is the variance.
On a high level, the variance can then be estimated by training separate same capacity models on different splits of the dataset, and then measuring the difference in outputs on the same test set. The bias is then computed by subtracting the empirical variance from the empirical risk.  For finer details, see \cite{yang2020rethinking}

For training, we follow \cite{yang2020rethinking} and train a ResNet-34 \citep{he2016deep} on CIFAR10, with stochastic gradient descent (learning rate $= 0.1$, momentum $= 0.9$). The learning rate is decayed a factor of 0.1 every 200 epochs, with a weight decay of 5e-4. The width $k$ of the network is varied suitably between 1 and 64. We also make 5 splits of 10,000 training samples for the calculation of bias/ variance.

\begin{figure}[ht]
  \centering
    \includegraphics[width=\linewidth]{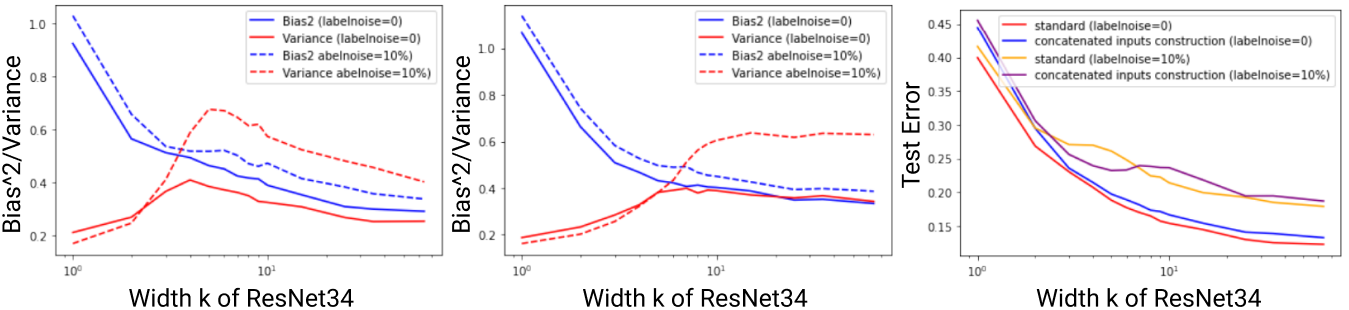}
    \label{fig:oneHotRethinking}\vspace{-0.6cm}
  \caption{
  Left: Bias and Variance against width k of ResNet-34 in the standard case. Middle: Bias and Variance against width k of ResNet-34 in the concatenated inputs construction. Right: Test error against width k of ResNet-34 in the standard case and concatenated inputs construction. %\textcolor{magenta}{Tasos: Try to include fontsize if possible.} I tried to find the original values but couldnt' find them.
  } \vspace{-0.8cm}
  \label{fig:RethinkingBiasVariancePlots}
\end{figure}

We present results in Figure \ref{fig:RethinkingBiasVariancePlots}. The concatenated inputs construction significantly delays and smoothens the increase in variance relative to the standard case, where the unimodal variance is significantly sharper. This impacts the shape of the test error, where in this setting we see a shifted bump in test error for the concatenated inputs construction. One possible explanation is in the case of deep neural networks the concatenated inputs construction is a form of implicit regularization for small models, which controls overfitting and leads to a smoother variance curve.
\vspace{-0.1cm}
\section{Discussion}
While the understanding of over/underparameterization and double descent is strongly tied to the number of samples, it appears that given a fixed number of unique samples it is possible to manipulate over/underparameterization by artificially boosting the number of samples by the concatenated inputs construction. 

A possible explanation in the literature for double descent is that the model is being forced to fit the training data as perfectly as possible, and at some model capacity it is possible to fit the training data perfectly by overfitting on non-existent, or weakly present, features. This results in overfitting and the double descent curve. Interestingly, the concatenated inputs construction generally mitigates double descent, even though it is possible to build models for the concatenated inputs construction from models for the standard setting. This suggests a possible route to improve the understanding of the relationship to the model capacity.

% \textbf{2. Double Descent Curve:} The existence of the double descent curve and the interpolation region is attributed to the number of parameters nearing the number of unique samples. One established explanation is the model is being forced to fit the training data as perfectly as possible, and at some model capacity it is possible to fit the training data perfectly by overfitting on non-existent, or weakly present, features. This results in overfitting and the double descent curve. However, the two-hot vector construction removes the double descent curve, even though it is possible to build models for the two-hot vector from models for the one-hot vector. This suggests an alternative or more nuanced explanation is more appropriate than just the model capacity.

%Consider underfitting: it is well known that overparameterized neural networks do not strongly overfit in practice \citep{zhang2016understanding}. 
The experimental results in this work support that neural network behavior can change with respect to the number of samples, even if the majority of samples add limited information, via the concatenated inputs construction. In this view, the concatenated inputs construction creates possibly a huge dataset, e.g. $50,000^2$ samples for the originally $50,000$ samples CIFAR10 where $50,000^2$ = 2,500,000,000 is far larger than any neural network for CIFAR10. Yet, there is no noticeable underfitting. Namely, the concatenated inputs construction quickly achieves comparable performance to the standard one-hot vector  deep learning setting. This suggests we may need to rethink the relationship between underfitting, the number of parameters, samples, and model capacity.

% \section{Conclusion}

% In this paper, we examine the double descent phenomena through the concatenated inputs construction. The construction is designed to add limited information whilst artificially augmenting the size of the dataset. As constructed, in the linear regression setting there is no impact on the location of the peak. Yet, in the neural network setting, the concatenated inputs construction generally mitigates the double descent curve. 
% % We also present observed results on the shifting of the double descent peak when moving from cross entropy to binary cross entropy loss. 
% Finally, we explore this phenomena through bias variance decomposition, and draw connections with samples, model capacity, and underfitting in discussion.

%\bibliographystyle{natbib}
% \bibliography{iclr2021_conference}

\bibliographystyle{ACM-Reference-Format}
\bibliography{iclr2021_conference}

% \appendix
% \section{Acceptable to have an appendix here}
% It is acceptable to include supplementary material here. 
% The main text has a four-page limit. References, acknowledgements and supplementary material can exceed the four pages. 

% Please submit a single PDF that contains both the main and the supplementary material. 
% \end{document}
%%%%%%%%%%%%%%%%%%%%%%%%%%%%%%%%%%%%%%%%%%%%%%%%%%%%%%%%%%%%%%%%%%%%%%%%%%%%%%%
%%%%%%%%%%%%%%%%%%%%%%%%%%%%%%%%%%%%%%%%%%%%%%%%%%%%%%%%%%%%%%%%%%%%%%%%%%%%%%%

\end{document}